\documentclass{article}
\usepackage{amsmath,epsfig}
\usepackage{epsfig}
\usepackage{amsmath}
\usepackage{amssymb}
\usepackage{color}
\usepackage{url}
\usepackage{multirow}
\usepackage{graphicx}
\usepackage{mathtools}
\usepackage{mathrsfs}
\usepackage{algorithm}
\usepackage{algorithmic}

\usepackage{bm}
\usepackage{graphicx}
\usepackage{bbm}
\usepackage{mathtools}
\usepackage{mathrsfs}

\usepackage{booktabs}
\usepackage{multirow}
\usepackage{algorithm}
\usepackage{algorithmic}
\usepackage{bbding}
\usepackage[table,xcdraw]{xcolor}
\usepackage{epsfig}
\usepackage{xcolor}
\usepackage{pifont}
\usepackage{amsmath}
\usepackage{subfigure}

\usepackage{hyperref}

\newlength\savewidth
\newcommand\shline{\noalign{\global\savewidth\arrayrulewidth
                            \global\arrayrulewidth 0.8pt}%
                   \hline
                   \noalign{\global\arrayrulewidth\savewidth}}
\usepackage{bbm}
\usepackage{bbding}
\usepackage[preprint]{spconfa4}

\let\OLDthebibliography\thebibliography
\renewcommand\thebibliography[1]{
  \OLDthebibliography{#1}
  \setlength{\parskip}{0pt}
  \setlength{\itemsep}{0pt plus 0.3ex}
}

\begin{document}\sloppy

% Example definitions.
% --------------------
\def\x{{\mathbf x}}
\def\L{{\cal L}}

% Title.
% ------
\title{ISTA-Net$^{++}$: Flexible Deep Unfolding Network for Compressive Sensing}
%
% Address.
% ---------------
\name{Di You$^{\ast}$,~Jingfen Xie$^{\ast}$,~Jian Zhang$^{\dag}$}
\address {Peking University Shenzhen Graduate School, Shenzhen, China}

\maketitle
\let\thefootnote\relax\footnotetext{$^{\ast}$ Equal contribution. $^{\dag}$ Corresponding author.}

\begin{abstract}
While deep neural networks have achieved impressive success in image compressive sensing (CS), most of them lack flexibility when dealing with multi-ratio tasks and multi-scene images in practical applications. To tackle these challenges, we propose a novel end-to-end flexible ISTA-unfolding deep network, dubbed ISTA-Net$^{++}$, with superior performance and strong flexibility. Specifically, by developing a dynamic unfolding strategy, our model enjoys the adaptability of handling CS problems with different ratios, i.e., multi-ratio tasks, through a single model. A cross-block strategy is further utilized to reduce blocking artifacts and enhance the CS recovery quality. Furthermore, we adopt a balanced dataset for training, which brings more robustness when reconstructing images of multiple scenes. Extensive experiments on four datasets show that ISTA-Net$^{++}$ achieves state-of-the-art results in terms of both quantitative metrics and visual quality. Considering its flexibility, effectiveness and practicability, our model is expected to serve as a suitable baseline in future CS research. The source code is available on \href{https://github.com/jianzhangcs/ISTA-Netpp}{https://github.com/jianzhangcs/ISTA-Netpp}.
\end{abstract}
\section{Introduction}
As a classical inverse problem, compressive sensing (CS) aims to recover an unknown signal from a small number of its measurements acquired by a linear random projection \cite{zhang2014group,zhang2018ista,zhao2018cream1}. Mathematically, suppose that  ${\mathbf x} \in \mathbb{R}^N$ is the original vectorized image block and $\mathbf{\Phi}\in \mathbb{R}^{M \times N}$ is a sampling matrix, the CS measurement of ${\mathbf x}$, denoted by ${\mathbf y}\in \mathbb{R}^M$ is usually formulated as $ \mathbf{y = \Phi x}$. The purpose of CS is to infer ${\mathbf x}$ from its randomized CS measurement ${\mathbf y}$. This inverse problem is typically ill-posed ($M \ll N$), whereby the CS ratio, denoted by $\gamma$, is defined as $\gamma=\frac{M}{N}$. 

\begin{figure}[t]
%\tiny
%\scriptsize
%\footnotesize
\centering
\includegraphics[width=1\linewidth]{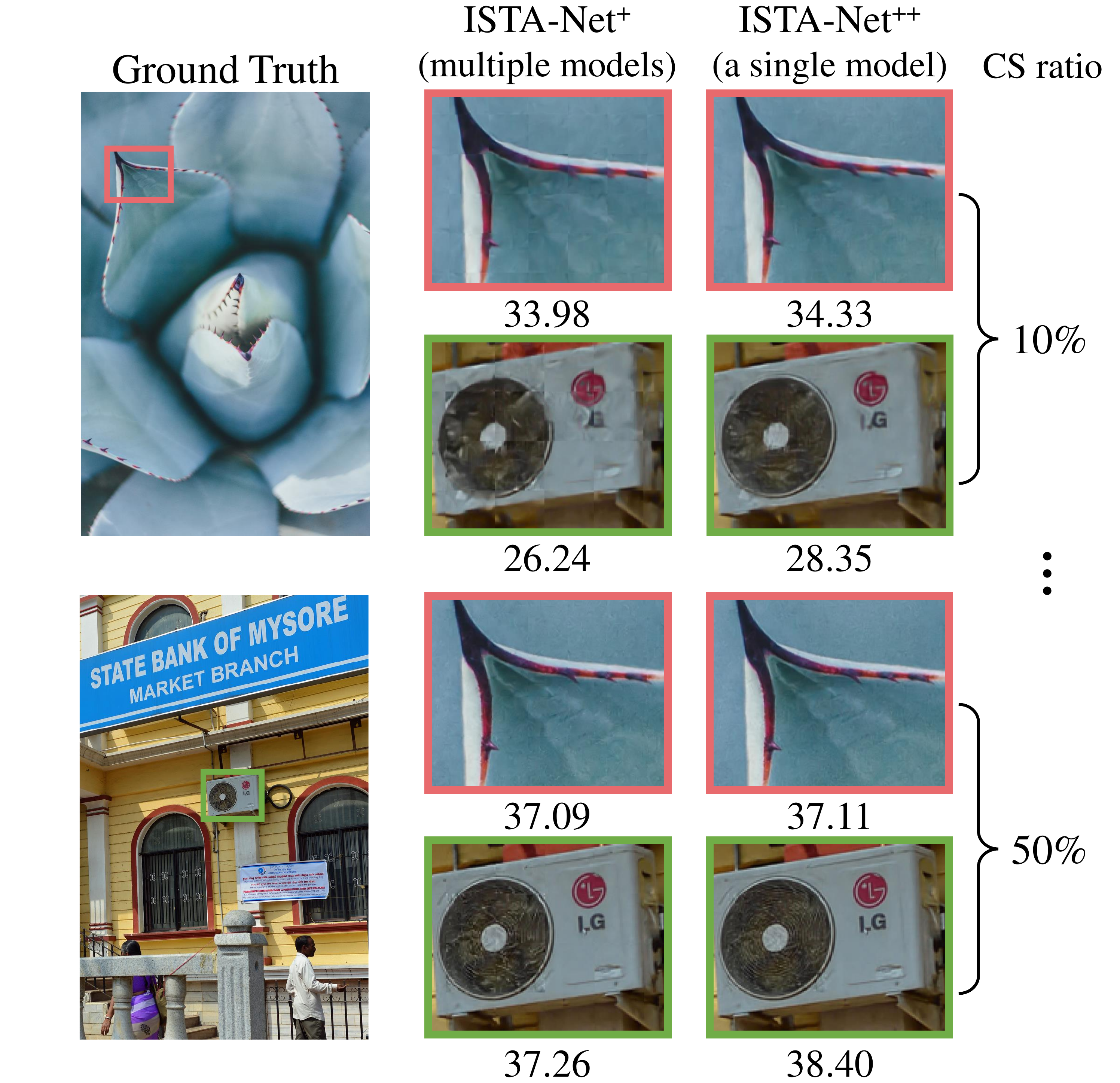}
\vspace{-20pt}
\caption{Visual and PSNR (dB) comparisons of ISTA-Net$^+$ and ISTA-Net$^{++}$. Our method not only handles multi-ratio tasks through a single model, but also overcomes the unbalance of the recovery quality in multiple scenes (e.g., ISTA-Net$^+$ behaves better in a plant scene but worse in a city scene).}
\label{fig: BSD68_fig1}  
\vspace{-10pt}
\end{figure}
%%%%%%%%%%%%%%%%%%%%%%%%%%%%%%%%%%%%%%%
Given the input linear measurements $\mathbf{y}$, traditional model-based CS methods \cite{zhang2012image,zhang2014imageSP} usually reconstruct the original image $\mathbf{{x}}$ by solving the following optimization problem:
\begin{equation}
\underset{{\mathbf{{x}}}}{\min} \frac{1}{2}\|\mathbf{\Phi \mathbf{{x}}- \mathbf{y}}\|^2_2 + \lambda{\psi({\mathbf{{x}})}},
\label{eq: classical CS}
\end{equation}
where $\psi(\mathbf{{x}})$ denotes the hand-crafted image prior term with $\lambda$ being the regularization parameter (generally pre-defined). A popular first-order proximal method is the iterative shrinkage-thresholding algorithm (ISTA) \cite{beck2009fast}, which solves the CS problem in Eq.~\eqref{eq: classical CS} by iterating between the following update steps:
%%%%%%%%%%%%%%%%%%%%%%%%%%%%%%%%%%%%%%%%%%
\vspace{-6pt}
\begin{equation}
% \footnotesize
\vspace{-5pt}
\label{eq: r}
\mathbf{r}_{k} = \mathbf{\hat{x}}_{k-1} - \rho \mathbf{\Phi}^{\top} (\mathbf{\Phi} \mathbf{\hat{x}}_{k-1} - \mathbf{{y}}),
\end{equation}
\vspace{-8pt}
\begin{equation}
\vspace{-5pt}
% \footnotesize
\label{eq: x}
\mathbf{\hat{x}}_{k} =  \underset{\mathbf{x}}{\arg\min}~\frac{1}{2}\|\mathbf{{x}} - \mathbf{r}_{k}\|^2_2  + \lambda{\psi({\mathbf{{x}})}},
\end{equation}
%%%%%%%%%%%%%%%%%%%%%%%%%%%%%%%%%%%%%%%%%%
where $k$ denotes the iteration index, and $\rho$ is the step size. 

Fueled by the rise of deep learning, data-driven neural networks with diverse modules \cite{shi2017deep,kulkarni2016reconnet,chenlearning,sun2020dual} have been proposed for image CS reconstruction by directly learning the inverse mapping from the CS measurement domain to the original signal domain. Most recently, some deep unfolding networks \cite{zhang2018ista, ren2019simultaneous, borgerding2017amp, yang2016deep, DBLP:journals/pami/DongWYSWL19,gilton2019neumann,zhang2020optimization} are developed to combine the merits of both the model- and data-driven methods and yield a better signal recovery performance. For example, one state-of-the-art method ISTA-Net$^+$ \cite{zhang2018ista} unfolds the previous ISTA update steps to a network consisting of a fixed number of layers, each of which corresponds to one iteration in traditional ISTA.

\begin{figure*}[t]
%\tiny
%\scriptsize
%\footnotesize
\centering
\includegraphics[width=1.0\linewidth]{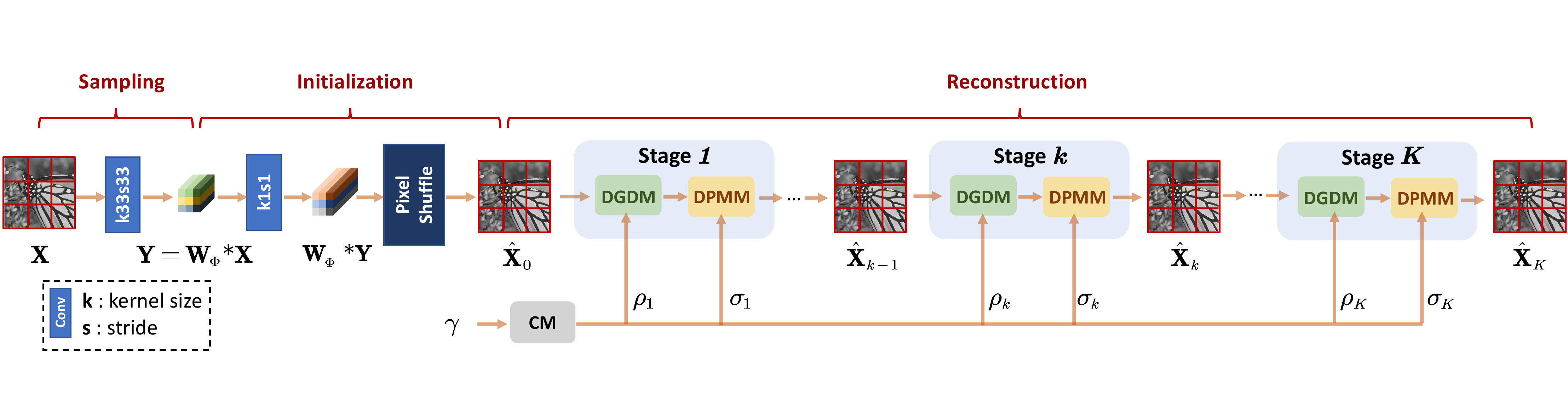}
%\vspace{-3mm}
\vspace{-35pt}
\caption{Illustration of our ISTA-Net$^{++}$ framework. Specifically, ISTA-Net$^{++}$ is composed of three parts: sampling, initialization, and reconstruction. In sampling process, images are measured using a sampling matrix which is considered as a convolution layer. The initialization process is to deal with the dimensionality mismatch between the original image and its CS measurement. And the reconstruction process is established by a deep unfolding model, which alternates between a dynamic gradient descent module (DGDM) and a dynamic proximal mapping module (DPMM), to generate the reconstruction image.}
\label{fig: framework}  
\vspace{-6pt}
\end{figure*}
%%%%%%%%%%%%%%%%%%%%%%%%%%%%%%%%%%%%%%%

However, most of the existing CS networks are not flexible enough to handle multi-ratio tasks and reconstruct multi-scene images in practice (see Fig.~\ref{fig: BSD68_fig1}). On one hand, different CS ratios are often considered as different independent tasks. As a result, for different CS ratios, these methods usually need to train multiple network models, resulting in large storage space and high time complexity, which does not satisfy the needs of real scenarios that usually contain various CS ratios.
On the other hand, we find that most CS networks suffer from unbalanced reconstruction results in various scenes. The reason is that their training dataset Train91 \cite{kulkarni2016reconnet} is unbalanced in the quantity distribution of different scenes, thus leading to undesirable results when reconstructing images of `rarely seen' scenarios \cite{kulkarni2016reconnet,chenlearning, sun2020dual,zhang2018ista}. 

To handle the above issues and further enhance the reconstruction quality, we propose a flexible deep unfolding network named ISTA-Net$^{++}$. Specifically, we unfold the traditional ISTA algorithm with a proposed dynamic unfolding strategy (DUS), which takes the CS ratio as an input and uses a condition module to transmit information about the input ratio to each stage, thus enabling our model to be trained under multiple ratios. Then, a cross-block strategy (CBS) is introduced to alleviate blocking artifacts and further improve the performance. Furthermore, to enable our model to handle images of various scenes more robustly, we adopt a more balanced training dataset and expect it to serve as a standard dataset in future CS research.

Overall, the contributions of this paper are four-fold: \textbf{1)} A flexible deep unfolding network named ISTA-Net$^{++}$ is proposed and enjoys much flexibility when handling multi-ratio tasks and multi-scene images in practical applications. \textbf{2)} ISTA-Net$^{++}$ develops a dynamic unfolding strategy and a cross-block strategy to promote network performance and adaptability, which enables our method to reconstruct CS images with different ratios through a single model. \textbf{3)} ISTA-Net$^{++}$ adopts a balanced training dataset to improve robustness when handling images of various scenarios. \textbf{4)} Experiments show that our approach performs favorably against state-of-the-arts in terms of flexibility, effectiveness and practicability and thus is expected to serve as a suitable baseline in future research.
\vspace{-10pt}
\section{Approach}
\vspace{-5pt}
\subsection{\textbf{Architecture Design of ISTA-Net$^{++}$}}
%%%%  
Similar to previous works \cite{kulkarni2016reconnet, zhang2018ista}, we propose to implement our ISTA-Net$^{++}$ in three parts: sampling, initialization, and reconstruction.
Concretely, as shown in Fig.~\ref{fig: framework}, the first part is to simulate the sampling process and obtain the randomized CS measurement. And the second part is to deal with dimensionality mismatch between original image and its CS measurement. Finally, the third part is to design the recovery network to generate the reconstructed image.

To address the issue of blocking artifacts \cite{zhang2021amp} coming from block-based sampling and reconstruction, we introduce the cross-block strategy (CBS) in our ISTA-Net$^{++}$. Since only using intra-block information to reconstruct a block results in blocking artifacts, we use the whole image $\mathbf{X} \in \mathbb{R}^{{H} \times {W}}$ instead of one vectorized image block $\mathbf{x}\in \mathbb{R}^N$ in our ISTA-Net$^{++}$, where $H$ and $W$ are multiples of $\sqrt{N}$. 

\vspace{-10pt}
\subsubsection{\textbf{Sampling and Initialization}}
To implement CBS in sampling process, we mimic the block-wise sampling process $\mathbf{y} = \mathbf{\Phi}\mathbf{x}$ equivalently by using a convolutional layer without bias and extend it to the whole image. As shown in Fig.~\ref{fig: framework}, the sampling process can be expressed as
% $\mathcal{A}(\MATHBF{X})=$
% $\mathbf{W}_{\Phi}*\mathbf{X}$
\vspace{-10pt}
\begin{equation}
\vspace{-5pt}
\mathbf{Y}=\mathcal{A}(\mathbf{X})=\mathbf{W}_{\Phi}*\mathbf{X},
\label{eq:sampling}
\end{equation}
% \vspace{-10pt}
where $*$ denotes the convolution operation.
To obtain $\mathbf{W}_{\Phi}$, we reshape the fixed random Gaussian matrix $\mathbf{\Phi} \in \mathbb{R}^{M \times N}$ into $M$ filters, each of which is of kernel size $ \sqrt{N} \times \sqrt{N}\times 1$. 
%%%%%%%%%%%%%%%%%%%%%%%%%%%%%%%%%%%%%%%%%%%

Correspondingly, the block-wise initialization $\mathbf{\hat{x}}^{(0)} = \mathbf{\Phi}^{\top}\mathbf{y}$ is implemented by a convolution layer followed by a pixel shuffle layer, and further extended to the whole image, which is defined as:
%%%%%%%%%%%%%%%%%%%%%%%%%%%%%%%%%%%%%%%%%%%
\vspace{-3pt}
\begin{equation}
\vspace{-3pt}
\mathbf{\hat{X}}_{0}=\mathcal{A}^{\top}(\mathbf{Y})=\operatorname{PixelShuffle}(\mathbf{W}_{{\Phi}^{\top}}*\mathbf{Y}).
\label{eq:initialization}
\end{equation}
%%%%%%%%%%%%%%%%%%%%%%%%%%%%%%%%%%%%%%%%%%%
Specifically, we first get $\mathbf{W}_{{\Phi}^{\top}}$ by reshaping $\mathbf{\Phi}^{\top} \in \mathbb{R}^{N \times M}$ into $N$ filters, each of which is of  kernel size $1\times 1\times M$. As a result, a $1\times 1$ convolution layer with weight $\mathbf{W}_{{\Phi}^{\top}}$ is utilized to obtain $\mathbf{\Phi}^{\top}\mathbf{y}$, which is actually a tensor of size $N\times 1\times 1$. Then, we adopt the pixel shuffle layer to reshape a tensor $N\times 1\times 1$ into a tensor $1\times \sqrt{N}\times \sqrt{N}$. Note that the design of our initialization process not only enables our network to deal with the dimension mismatch between the original image and its CS measurement, but also serves as a naive solution to handle sampling matrices with various dimensions (CS ratios).

\vspace{-10pt}
\subsubsection{\textbf{Reconstruction}}

As a deep unfolding network, the reconstruction process of our ISTA-Net$^{++}$ unfolds the traditional ISTA and alternates between a dynamic gradient descent module (DGDM) and a dynamic proximal mapping module (DPMM) for $K$ times (shown in Fig.~\ref{fig: framework}), corresponding to Eq.~\eqref{eq: r} and Eq.~\eqref{eq: x} respectively.
Specifically, to deal with multiple ratios and make our ISTA-Net$^{++}$ more flexible, we propose a dynamic unfolding strategy (DUS) to unfold ISTA, which takes the CS ratio as an input and uses a global condition module (CM) to transmit information about the input ratio to each stage.

%%%%%%%%%%%%%%%%%%%%%%%%%%%%%%%%%%%%%%%
\begin{figure}[t]
%\tiny
%\scriptsize
%\footnotesize
\centering
\includegraphics[width=1\linewidth]{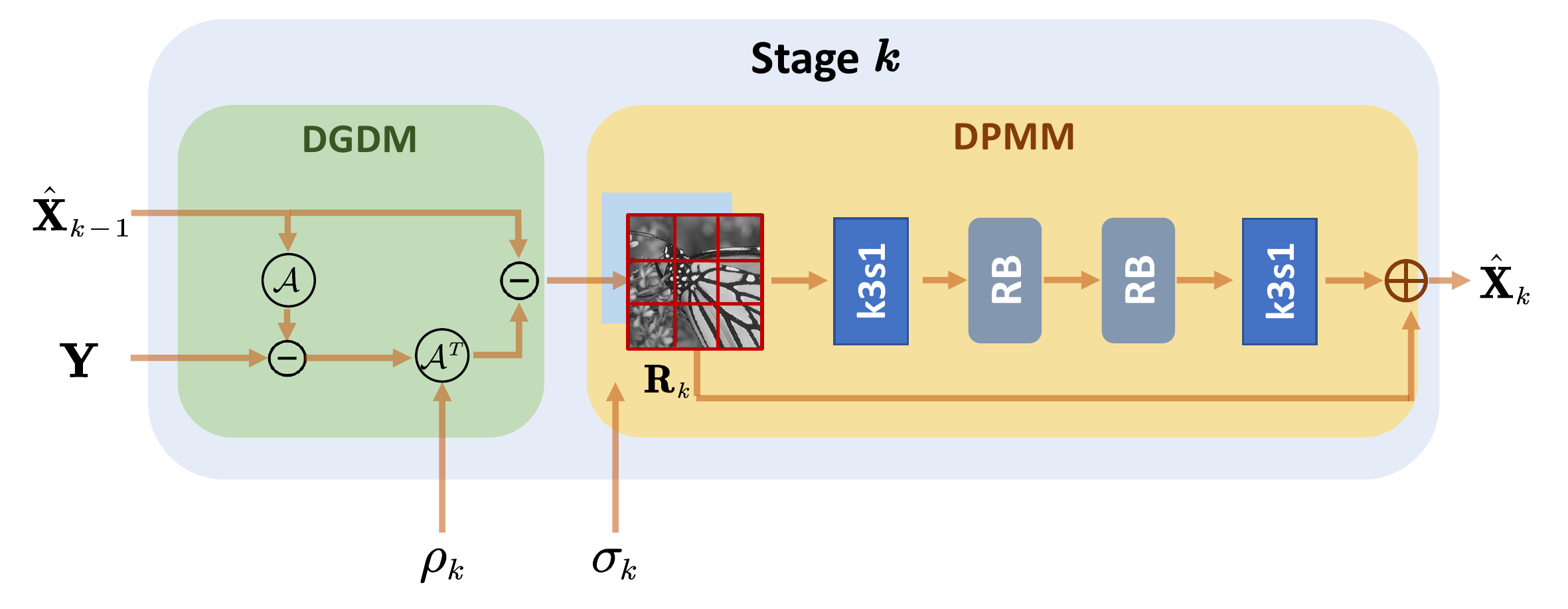}
%\vspace{-3mm}
\vspace{-26pt}
\caption{Illustration of the $k$-th stage of the reconstruction process in our ISTA-Net$^{++}$. Specifically, it is composed of DGDM and DPMM, controlled by $\rho_k$ and $\sigma_k$ respectively. }
\label{fig: framework2}  
\vspace{-5pt}
\end{figure}
%%%%%%%%%%%%%%%%%%%%%%%%%%%%%%%%%%%%%%%

\textbf{Dynamic Gradient Descent Module (DGDM):}
% \subsubsection{$\mathbf{r}^{{(k)}}$ \textbf{Module:}}
To preserve the ISTA structure, DGDM is directly defined according to Eq.~\eqref{eq: r} (see Fig.~\ref{fig: framework2}), in which $\rho_{k}$ comes from the output of CM.
So taking the $\mathbf{\hat{X}}_{k-1},\mathbf{\Phi},\mathbf{{Y}}$, and $\rho_{k}$ as input, the output of the DGDM in $k$-th stage is finally defined as:
%%%%%%%%%%%%%%%%%%%%%%%%%%%%%%%%%%%%%%%%
\vspace{-5pt}
\begin{equation}
\vspace{-8pt}
\begin{aligned}
\mathbf{R}_{k} &= \mathcal{H}^{DGDM}_{k}(\mathbf{\hat{X}}_{k-1},\mathbf{\Phi},\mathbf{{Y}},\rho_{k}) \\
&=\mathbf{\hat{X}}_{k-1} - \rho_{k} \mathcal{A}^T  (\mathcal{A}(\mathbf{\hat{X}}_{k-1}) - \mathbf{{Y}}),
\end{aligned}
\end{equation}
%%%%%%%%%%%%%%%%%%%%%%%%%%%%%%%%%%%%%%%%
where $\mathbf{\hat{X}}_{0}$ is the output of the initialization process. Though learning a fixed $\rho_{k}$ in DGDM achieves great results when tackling one fixed ratio, considering the need of handling various ratios, we argue that a performance gain can be obtained if $\rho_{k}$ varies with CS ratio $\gamma$, which influences the degree of ill-posedness.

%%%%%%%%%%%%%%%%%%%%%%%%%%%%%%%%%%%%
\begin{table*}[t]
\centering
\footnotesize
% \small
% \vspace{-5pt}
\setlength{\tabcolsep}{3pt}   
% \vspace{3pt}
\caption{Average PSNR performance comparisons of various CS methods with different CS ratios on the BSD68 and Set11 datasets. The best and second best results are highlighted in \textcolor[rgb]{1.00,0.00,0.00}{red} and \textcolor[rgb]{0.00,0.00,1.00}{blue} colors, respectively. $\times n$ represents $n$ trained models for different CS ratios. As we can see, our approach achieves the best results under all CS ratios with a single trained model. }
\begin{tabular}{l|l|cccccc|cccccc|c}
\shline
\multirow{3}{*}{Methods}                                                     & \multirow{3}{*}{Parameters} & \multicolumn{6}{c|}{Datasets: BSD68}          & \multicolumn{6}{c|}{Datasets: Set11}           & \multirow{3}{*}{\begin{tabular}[c]{@{}c@{}}Time\\ CPU/GPU\end{tabular}} \\ \cline{3-14}
                                                                             &                        & \multicolumn{6}{c|}{CS Ratio $\gamma$}                  & \multicolumn{6}{c|}{CS Ratio $\gamma$}                 &                                                                         \\ \cline{3-14}
                                                                             &                        & 10\%  & 20\%  & 30\%  & 40\%  & 50\%  & Avg.  & 10\%  & 20\%  & 30\%  & 40\%  & 50\%  & Avg.  &                                                                         \\ \hline
BM3D-AMP   \cite{metzler2016denoising}(TIT2016)             & --                     & 22.68 & 24.77 & 26.44 & 28.19 & 29.86 & 26.39 & 22.6  & 26.77 & 30.26 & 33.66 & 35.93 & 29.84 & 44.58s/-                                                                \\
ReconNet   \cite{kulkarni2016reconnet}(CVPR2016)          & \color{red} 0.14M $\times$ 5 = 0.70M                  & 23.88 & 25.75 & 26.72 & 28.96 & 30.13 & 27.09 & 24.06 & 26.68 & 28.14 & 30.78 & 31.48 & 28.23 & -/0.0011s                                                               \\
LDAMP   \cite{DBLP:conf/nips/MetzlerMB17}(NeurIPS2017)   & --                     & 23.94 & 27.74 & \color{blue}30.28 &\color{blue} 32.12 & 32.89 & 29.39 & 24.71 & 30.65 & 33.87 & 36.03 & 36.60  & 32.37 & -/390.40s                                                               \\
DIP   \cite{ulyanov2018deep}(CVPR2018)                    & --                 & 25.05 & 27.25 & 28.66 & 29.82 & 31.21 & 28.40 & 25.98 & 29.81 & 33.25 & 33.41 & 35.96 & 31.68 & -/335.87s                                                               \\
ISTA-Net$^+$   \cite{zhang2018ista}(CVPR2018)            & 0.34M $\times$ 5 = 1.70M                 & 25.24 & 28.00 & 30.20 & 32.10 & 33.93 &\color{blue} 29.89 & 26.57 & 30.85 & 33.74 & 36.05 &\color{blue} 38.05 & 33.05 & -/0.0051s                                                               \\
DPDNN   \cite{DBLP:journals/pami/DongWYSWL19}(TPAMI2019) & 1.36M $\times$ 5 = 6.80M                    & 24.81 & 27.28 & 29.22 & 30.99 & 32.74 & 29.01 & 26.09 & 29.75 & 32.37 & 34.69 & 36.83 & 31.95 & -/0.0951s                                                               \\
GDN \cite{gilton2019neumann} (TCI2019)                  & 0.50M $\times$ 5 = 2.50M                    & 25.19 & 27.95 & 29.88 & 32.07 & \color{blue}34.09 & 29.84 & 26.03 & 30.16 & 32.95 & 35.25 & 37.60  & 32.40 & -/0.0061s                                                               \\
NLR-CSNet \cite{8999514}   (TMM2020)                       & --                     & 25.23 & 27.69 & 29.55 & 31.14 & 32.57 & 29.24 & \color{blue}28.05 & \color{blue}31.64 & \color{blue}33.89 & 35.65 & 37.12 & 33.27 & -/378.75s                                                               \\
DPA-Net \cite{sun2020dual}  (TIP2020)                    & 9.31M $\times$ 5 = 46.5M                    & 25.33 & --    & 29.58 & --    & --    & --    & 27.66 & --      & 33.60 & --    & --    & --    & -/0.0365s                                                               \\
MAC-Net \cite{chenlearning}  (ECCV2020)                  & 6.12M $\times$ 5 = 30.6M                     & \color{blue}25.70 & \color{blue}28.23 & 30.10 & 31.89 & 33.37 & 29.86 & 27.92 & 31.54 & 33.87 & \color{blue}36.18 & 37.76 & \color{blue}33.45 & -/0.0710s                                                               \\
ISTA-Net$^{++}$ (Ours)                                  & \color{blue} 0.76M $\times$ 1 = 0.76M                  & \multicolumn{1}{c}{\color{red}{26.25}} & \multicolumn{1}{c}{\color{red}{29.00}} & \multicolumn{1}{c}{\color{red}{31.10}} & \multicolumn{1}{c}{\color{red}{33.00}} & \multicolumn{1}{c}{\color{red}34.85} & \multicolumn{1}{c|}{\color{red}30.84} &  \multicolumn{1}{c}{\color{red}{28.34}} & \multicolumn{1}{c}{\color{red}{32.33}} & \multicolumn{1}{c}{\color{red}{34.86}} & \multicolumn{1}{c}{\color{red}{36.94}} & \multicolumn{1}{c}{\color{red}{38.73}}&\multicolumn{1}{c|}{\color{red}{34.24}}& -/0.0123s                                                               \\ \shline
\end{tabular}
\label{tab:result_noise_0}
\vspace{-10pt}
\end{table*}
%%%%%%%%%%%%%%%%%%%%%%%%%%%%%%%%%%%%
\begin{table}[]
\caption{Average PSNR comparisons on high resolution images, i.e., Urban100 and DIV2K validation datasets.}
\setlength{\tabcolsep}{3pt}  
\label{tab:div2k_urban100}
\footnotesize
\begin{tabular}{l|cccc|cccc}
\shline
\multirow{3}{*}{Methods} & \multicolumn{4}{c|}{Datasets: Urban100} & \multicolumn{4}{c}{Datasets: DIV2K}   \\ \cline{2-9} 
                         & \multicolumn{4}{c|}{CS Ratio $\gamma$}           & \multicolumn{4}{c}{CS Ratio $\gamma$}          \\ \cline{2-9} 
                         & 10\%    & 30\%  & 50\%  & Avg.  & 10\%   & 30\%  & 50\%  &  Avg. \\ \hline
ISTA-Net$^{+}$                 & 23.62   & \color{blue}30.07  & \color{blue}34.39 &\color{blue}29.36 & 27.78 & \color{blue}33.37 & \color{blue}37.35 &\color{blue}32.83 \\
DPA-Net                  & \color{blue}24.55    & 29.47 & --   &  -- & \color{blue}28.23   & 33.04     & -- & --  \\
MAC-Net                  & 24.26    & 29.46 & 33.36 & 29.03 & 27.86 & 32.90 &  36.30 &32.35\\
ISTA-Net$^{++}$                 & \color{red}25.53    & \color{red}31.93  & \color{red}35.84 & \color{red}31.10 & \color{red}28.96  &\color{red} 34.41 &\color{red} 38.22 & \color{red}33.86\\ \shline
\end{tabular}
\vspace{-10pt}
\end{table}
%%%%%%%%%%%%%%%%%%%%%%%%%%%%%%%%%%%%%%%
\begin{figure}[t]
%\tiny
%\scriptsize
%\footnotesize
\centering
\includegraphics[width=0.9\linewidth]{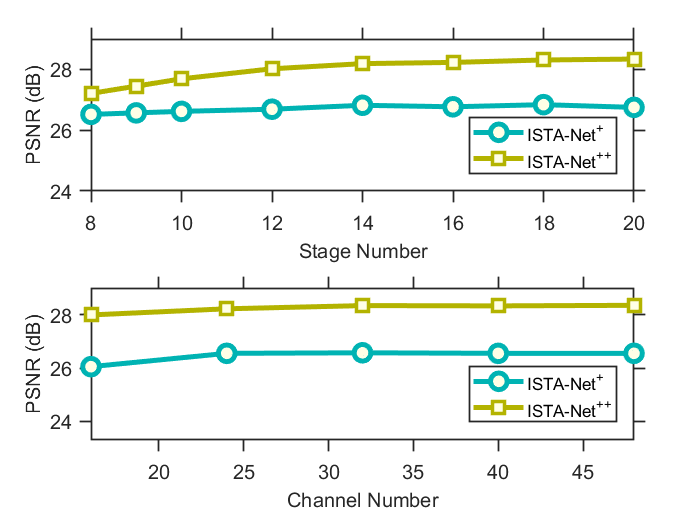}
%\vspace{-3mm}
\vspace{-15pt}
\caption{PSNR comparison between ISTA-Net$^{+}$ \cite{zhang2018ista} and ISTA-Net$^{++}$ with various numbers of stages and channels.}
\label{fig: compare with orgin}  
\vspace{-10pt}
\end{figure}
%%%%%%%%%%%%%%%%%%%%%%%%%%%%%%%%%%%%%%%
\textbf{Dynamic Proximal Mapping Module (DPMM):}
To solve the proximal mapping problem in Eq.~(\ref{eq: x}), we propose an effective and flexible module, DPMM. By taking an additional noise level map $\mathbf{M}^{\sigma}_{k}$ as input, we make the DPMM flexible to multiple degradation levels \cite{DBLP:journals/tip/ZhangZZ18}.
Specifically, filled with $\sigma_k$ coming from CM, the map $\mathbf{M}^{\sigma}_{k}$ has the same spatial size as the input image of DPMM. As shown in Fig.~\ref{fig: framework2}, DPMM in the $k$-th stage, denoted as $\mathcal{H}_{k}^{DPMM}$, can be formulated as:
%%%%%%%%%%%%%%%%%%%%%%%%%%%%%%
\begin{equation}
\vspace{-2pt}
\begin{aligned}
\mathbf{\hat{X}}_{k} &=\mathcal{H}_{k}^{DPMM}(\mathbf{R}_{k},\sigma_{k}) \\ &=\mathbf{R}_{k}+ \mathcal{H}^{rec}_{k}(\mathcal{H}_{k}^{RB,2}({\mathcal{H}_{k}^{RB,1}}(\mathcal{H}^{ext}_{k}(|\mathbf{R}_{k},\mathbf{M}^{\sigma}_{k}|)))).
\end{aligned}
\vspace{-2pt}
\end{equation}
Here, DPMM is composed of 2 residual blocks (RBs) $\mathcal{H}_{k}^{RB,1}$ and $\mathcal{H}_{k}^{RB,2}$, two convolution layers $\mathcal{H}^{ext}_{k}$, $\mathcal{H}^{rec}_{k}$, which devote to extracting the image features and reconstruction, respectively, and a long skip connection.

\textbf{Condition Module (CM):}
To deal with multiple ratios with a single model more flexibly, Condition Module (CM) is designed to predict the condition information transmitted to each stage. 
The condition information transmitted to DGDM and DPMM ($\rho_{k}$ and $\sigma_{k}$) has great potential in promoting the generalization performance. 
Let $\mathbf{F}^{CM}_{k} =[\rho_{k},\sigma_{k}]$ , the condition information generated by CM can be expressed as
%%%%%%%%%%%%%%%%%%%%%%%%%%%%%%%%%%%%%%%%%%
\vspace{-1pt}
\begin{equation}
\vspace{-1pt}
\mathbf{F}^{CM}=[\mathbf{F}^{CM}_{1},\mathbf{F}^{CM}_{2},\dots,\mathbf{F}^{CM}_{K}] =\mathcal{H}^{CM}(\gamma).
\end{equation}
%%%%%%%%%%%%%%%%%%%%%%%%%%%%%%%%%%%%%%%%
% Inspired by \cite{DBLP:conf/cvpr/ZhangGT20}, the CM consists of three fully. 
Through preliminary experiments, we adopt three fully connected layers with ReLU as the first two activation functions and Softplus as the last to implement CM, similar to \cite{DBLP:conf/cvpr/ZhangGT20}.

\vspace{-10pt}
\subsection{\textbf{Network Parameters and Loss Function}}
First, given the training dataset $\left \{ \mathbf{X}^i \right \}_{i=1}^{N_D}$ and the sampling matrix set $\{\mathbf{\Phi}^t\}_{t=1}^{N_{\gamma}}$, we get $\mathbf{Y}^{i,t}$. Then, taking $\mathbf{Y}^{i,t}, \mathbf{\Phi}^t, \gamma^t$ as input, ISTA-Net$^{++}$ aims to reduce the discrepancy between $\mathbf{X}^i$ and the result of reconstruction process $\mathcal{H}(\mathbf{Y}^{i,t}, \mathbf{\Phi}^t,\gamma^t)$. Therefore, we design the loss function to train ISTA-Net$^{++}$:
%%%%%%%%%%%%%%%%%%%%%%%%%%%
%%%%%%%%%%%%%%%%%%%%%%%%%%%
\begin{equation}
% \footnotesize
\label{eq: LossFunction}
% \small
{\mathcal{L}}(\mathbf{\Theta}) =  \frac{{\sum^{N_D}_{i=1}\sum^{N_{\gamma}}_{t=1}\|\mathcal{H}(\mathbf{Y}^{i,t},\mathbf{\Phi}^t,\gamma^t) - \mathbf{X}^i\|^2_2}}{N_DN_{\gamma}},
\end{equation}
where $\mathbf{\Theta}$ denotes the learnable parameter set in ISTA-Net$^{++}$, including the parameters of the CM, DGDM and DPMM, i.e., $\mathcal{H}^{CM}(\cdot)$, $\mathcal{H}_{k}^{DGDM}(\cdot)$, $\mathcal{H}_{k}^{DPMM}(\cdot)$

\vspace{-10pt}
\section{Experiments}
\vspace{-5pt}
\subsection{\textbf{Implementation Details}}
\begin{figure}[t]
%\tiny
%\scriptsize
%\footnotesize
\centering
\includegraphics[width=1.0\linewidth]{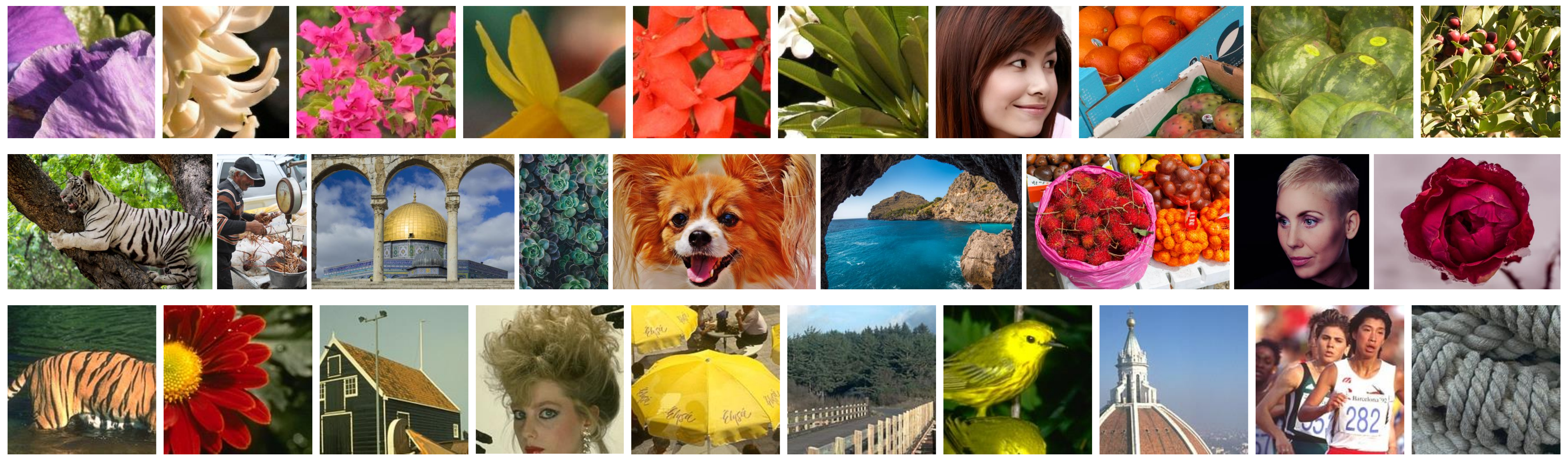}
\vspace{-18pt}
\caption{From top to bottom, each row shows some representative images in Train91, DIV2K, Train400 datasets, respectively. Obviously, most images in Train91 are actually plants with similar scenes, lacking diversity.}
\label{fig: 3datasets}  
\vspace{-10pt}
\end{figure}

Following \cite{kulkarni2016reconnet}, we use 400 images of size 180$\times$180 for training (see the third row in Fig.~\ref{fig: 3datasets}). As for testing, we utilize four widely-used benchmark datasets: Set11 \cite{kulkarni2016reconnet}, BSD68 \cite{martin2001database}, Urban100 \cite{Huang-CVPR-2015}, and the validation dataset in DIV2K \cite{timofte2017ntire}. To train the network, we use Adam \cite{kingma2015adam} with a batch size of 64 and $K$ is set to 20. Note that the CS recovered results are evaluated with Peak Signal-to-Noise Ratio (PSNR).

\vspace{-8pt}
\subsection{ISTA-Net$^+$ vs. ISTA-Net$^{++}$}
To demonstrate the superiority of ISTA-Net$^{++}$ over the classic deep unfolding method ISTA-Net$^+$, we compare them in various numbers of stages and channels. Fig.~\ref{fig: compare with orgin} shows the average PSNR curves on Set11 with respect to different stage numbers and channel numbers, when the CS ratio is 10$\%$. While the PSNR performance of both methods gets higher with the increase of stage number and channel number, ISTA-Net$^{++}$ achieves better recovery performance with fewer parameters. For example, ISTA-Net$^++$ achieves 0.88/1.46 dB gain over ISTA-Net$^+$ when K=9/20. Note that our ISTA-Net $^{++}$ achieves excellent performance while handling multiple CS ratios with a single model. Considering the tradeoff between network complexity and recovery performance, we set the default stage number to be 20 and channel number to be 32 for ISTA-Net$^{++}$ in the following experiments. 
%%%%%%%%%%%%%%%%%%%%%%%%%%%
%%%%%%%%%%%%%%%%%%%%%%%%%%%%%%%%%%%%%%

\begin{figure*}[t]
%\tiny
%\scriptsize
%\footnotesize
\centering
\includegraphics[width=1\linewidth]{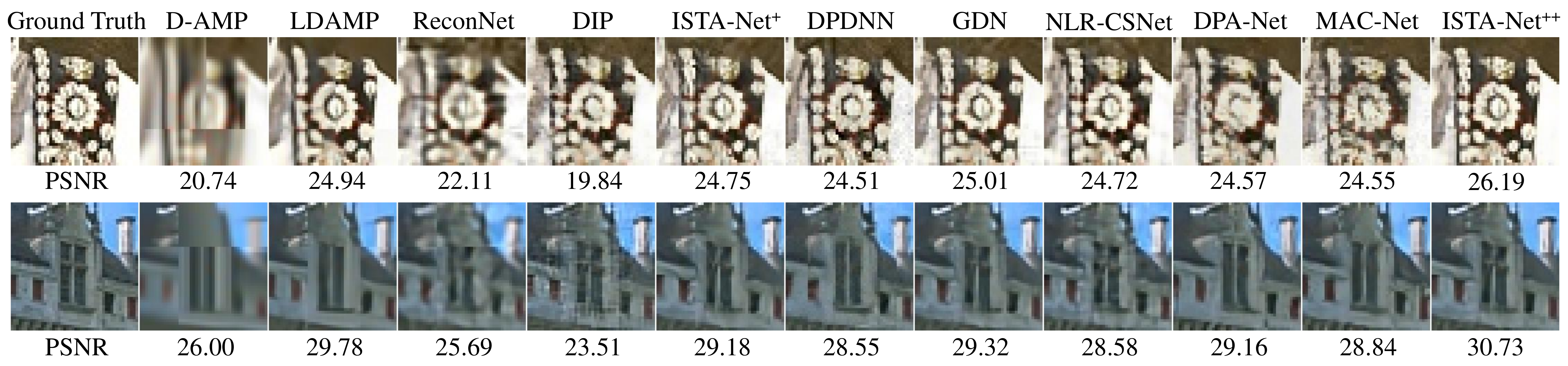}
\vspace{-23pt}
\caption{Visual Result of different methods on BSD68 dataset (CS ratio is 30\%).}
\label{fig: bsd68}  
\vspace{-10pt}
\end{figure*}
%%%%%%%%%%%88%%%%%%%%%%%%%%%%%%

\vspace{-13pt}
\subsection{\textbf{Comparison with State-of-the-Art Methods}}
\vspace{-2pt}

%%%%%%%%%%%%%%%
\begin{figure}[t]
%\tiny
%\scriptsize
%\footnotesize
\centering
\includegraphics[width=0.95\linewidth]{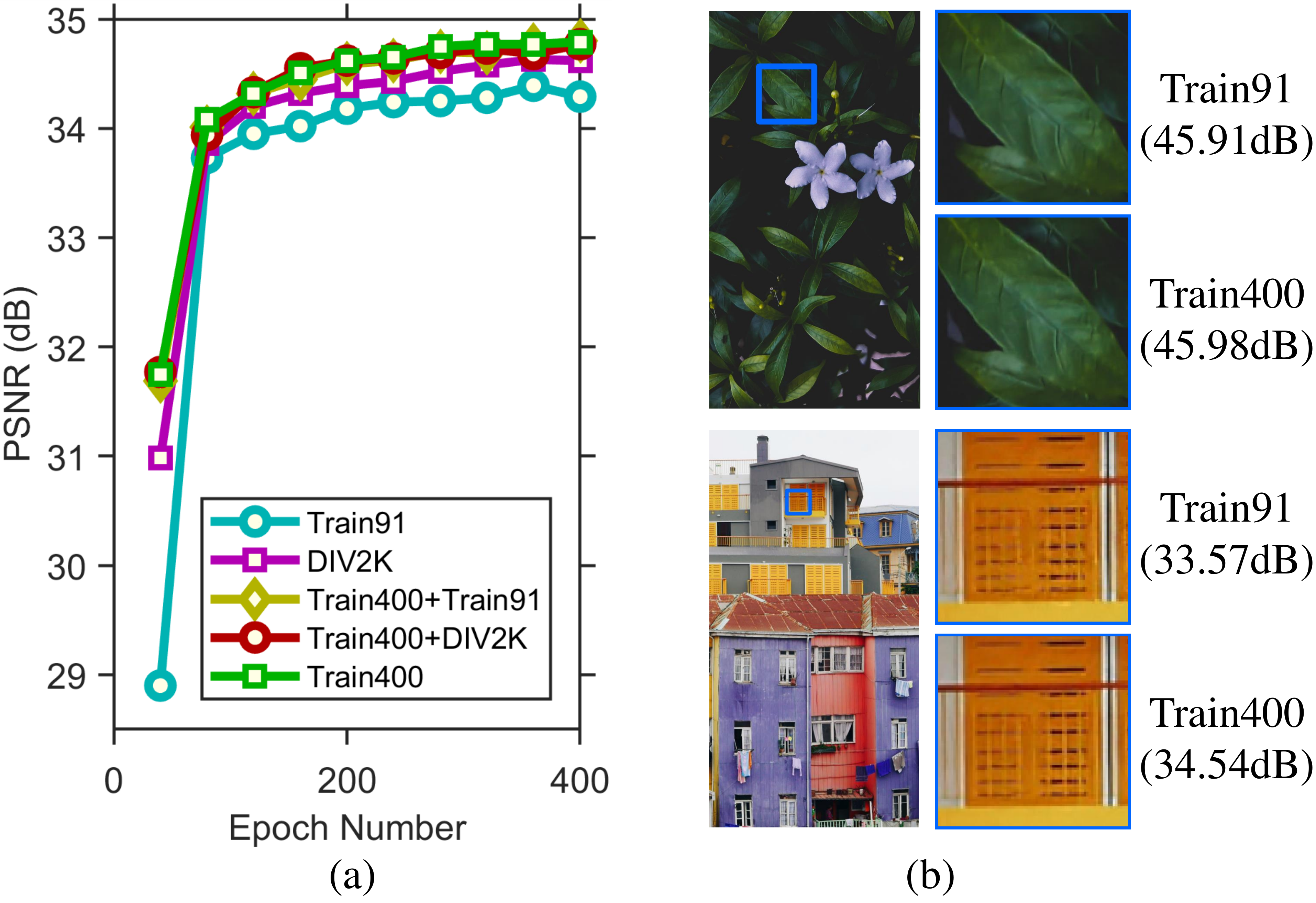}
\vspace{-10pt}
\caption{(a) PSNR (dB) comparison among five versions of ISTA-Net$^{++}$ (trained on Train91, Train400, DIV2K, Train400+Train91, and Train400+DIV2K). (b) Visual results of multi-scene recovery of ISTA-Net$^{++}$ trained on two different datasets. ISTA-Net$^{++}$ trained on Train91 behaves unbalanced, e.g., better in a plant scene but worse in a city scene.}
\label{fig: epoch_psnr}  
\vspace{-10pt}
\end{figure}
%%%%%%%%%%%88%%%%%%%%%%%%%%%%%%

Our proposed ISTA-Net$^{++}$ is compared with several representative state-of-the-art methods including BM3D-AMP \cite{metzler2016denoising}, LDAMP \cite{DBLP:conf/nips/MetzlerMB17}, DIP \cite{ulyanov2018deep}, ReconNet \cite{lohit2018convolutional}, DPDNN \cite{DBLP:journals/pami/DongWYSWL19}, GDN \cite{gilton2019neumann}, ISTA-Net$^+$ \cite{zhang2018ista}, NLR-CSNet \cite{8999514}, DPA-Net \cite{sun2020dual}, and MAC-Net \cite{chenlearning}.
Different from other deep network-based end-to-end methods, for ISTA-Net$^{++}$, we use the five sampling matrices for five CS ratios to train our model only once. 

Table~\ref{tab:result_noise_0} and Table~\ref{tab:div2k_urban100} list the average PSNR results for different ratios and parameters for various CS methods. We can observe that our ISTA-Net$^{++}$ achieves the highest PSNR results. For example, our ISTA-Net$^{++}$ achieves on average 0.95/1.19/1.74/1.03  dB PSNR gains over the state-of-the-art method ISTA-Net$^{+}$ on BSD68/Set11/Urban100/DIV2K validation dataset, respectively. Furthermore, for five different CS ratios, other deep network-based end-to-end CS methods need to train five different network models separately, resulting in five times of parameters storage, which can be avoided in our ISTA-Net$^{++}$.
In addition, Table~\ref{tab:result_noise_0} shows the average running time (in seconds) of various algorithms for reconstructing a 256$\times$256 image, which clearly indicates that our ISTA-Net$^{++}$ produce consistently better reconstruction results, while remaining computationally attractive real-time speed. 
Fig.~\ref{fig: bsd68} shows the visual comparisons of all the competing methods when CS ratio is 10\% on BSD68 dataset. It can be seen that our ISTA-Net$^{++}$ has much higher PSNR than other methods and is able to recover more details and much sharper edges than other competing methods.

\vspace{-8pt}
\section{Analysis}
\vspace{-5pt}
\subsection{Analysis on Robustness for Various Scenes} 
\vspace{-5pt}
Following \cite{kulkarni2016reconnet}, most of the current state-of-the-art CS networks \cite{kulkarni2016reconnet,zhang2018ista,sun2020dual,chenlearning} use a unified training dataset, i.e., Train91 (see the first row in Fig.~\ref{fig: 3datasets}), for the convenience of comparison. 
However, 78\% images in Train91 dataset are plants with similar scenes, which obviously does not satisfy the need of the real-world CS problem that usually contains a large number of various scenarios. 
Therefore, we select another two datasets (DIV2K training dataset, Train400) commonly used in low-level vision to explore a more suitable dataset for current CS models  (see the second and third rows in Fig.~\ref{fig: 3datasets}). The test convergence curves of using different training datasets are shown in Fig.~\ref{fig: epoch_psnr}(a).
As can be observed, with the same number of epochs (i.e.,400), the model trained on Train400 achieves 34.81 dB, higher than 34.25 dB yielded by that trained on Train91. Compared with the model trained on Train400, using a larger training dataset (Train400+Train91, Train400+DIV2K) can only bring negligible improvements, showing that the balance of the image scenes has a greater impact on the performance gain than the size of training dataset.
Furthermore, the results on other existing methods shown in Table~\ref{tab:train400} verify the superiority of the Train400 dataset.

In order to further verify the improvement on the robustness for various scenes, we compare the multi-scene reconstruction performances of the two models separately trained on Train400 and Train91. Fig.~\ref{fig: epoch_psnr}(b) shows the visual comparisons of the two versions of our model when the CS ratio is 30\% on DIV2K validation dataset. For the plant scene which appears a lot in Train91 (see the first example in Fig.~\ref{fig: epoch_psnr}(b)), the model trained on Train91 and Train400 both achieve desirable visual performance and similar PSNR results. 
However, as for the house scenario which is `rarely seen' by Train91, ISTA-Net$^{++}$ trained on Train91 behaves 0.97dB worse than that  trained on Train400 (see the second example in Fig.~\ref{fig: epoch_psnr}(b)), which proves the imbalance of the image scenes of Train91 and further verifies the robustness brought by adopting the more balanced Train400 dataset.
\vspace{-5pt}
\begin{table}[t]
\vspace{-5pt}
\centering
\caption{Average PSNR performance comparisons of various CS methods trained on different datasets(CS ratio is 30\% and the test dataset is BSD68). Thanks to multi-scene training samples, network trained on Train400 achieves a performance gain compared with that trained on Train91.}
\vspace{2pt}
\label{tab:train400}
\footnotesize
\setlength{\tabcolsep}{10pt} 
\begin{tabular}{c|ccc}
\shline
Training Dataset & ISTA-Net$^{+}$   & DPDNN & MAC-Net \\ \hline
Train91  & 30.20     & 29.22 & 30.10   \\
Train400 & 30.68     & 30.13 & 30.68   \\ \shline
\end{tabular}
\vspace{-9pt}
\end{table}

\vspace{-5pt}
\subsection{Ablation Study and Discussion}
\vspace{-5pt}
In order to better understand the behaviour of ISTA-Net$^{++}$, we conduct two groups of ablation studies to evaluate the influence of DUS and CBS on the reconstruction performance.
%%%%%%%%%%%%%%%%%%%%%%%%%%%%%%%%%%%%%%%%

\textbf{Effectiveness of DUS:} By adopting a condition module, our ISTA-Net$^{+}$ enjoys the flexibility of handling CS problems with different ratios through a single model.
We remove the additional CM and directly train the network with the same five sampling matrices with our ISTA-Net$^{++}$.
Settings (b) and (e) in Table~\ref{tab:ablation_study} provide the performance comparison for three CS ratios on Set11, which shows that ISTA-Net$^{++}$ with DUS consistently outperforms the one without CM across all the ratios and obtains on average 0.26 dB gains. And setting (b) and (c) further prove the influence of the two parts of DUS.
\vspace{-5pt}
\begin{table}[t]
\centering
\footnotesize 
\caption{Ablation study of different components, which evaluates the effectiveness of our proposed DUS and CBS. } 
\label{tab:ablation_study}
\setlength{\tabcolsep}{3.2pt} 
\begin{tabular}{c|c|c|c|c|ccc}
\shline
\multicolumn{1}{c|}{\multirow{2}{*}{Setting}} &
\multicolumn{1}{c|}{\multirow{2}{*}{DUS-$\rho$}} &
\multicolumn{1}{c|}{\multirow{2}{*}{DUS-$\sigma$}} &
\multicolumn{1}{c|}{\multirow{2}{*}{CBS}}  & \multicolumn{1}{c|}{\multirow{2}{*}{Parameters}} & \multicolumn{3}{c}{CS Ratio $\gamma$}  
\\ \cline{6-8} \multicolumn{1}{c|}{}   & \multicolumn{1}{c|}{} & \multicolumn{1}{c|}{} & \multicolumn{1}{c|}{} & \multicolumn{1}{c|}{}   & 10\% &  30\%  & 50\% \\ \hline
(a)      & \ding{55}  & \ding{55} & \ding{55}          &   752,040         &  27.14         &  34.06 & 38.02           \\ 
(b)      & \ding{55}  & \ding{55} & \ding{51}        &   752,040         &   28.12         &  34.65    &38.37       \\ 
(c)    &        \ding{51}  & \ding{55}   & \ding{51}       &752,040&   28.18      &  34.75    &   38.48                \\ 
(d)   & \ding{55}     & \ding{51}        & \ding{51}      &759,580& 28.17       & 34.79       & 38.64            \\ 
(e)      & \ding{51}   & \ding{51}    &\ding{51}        & 760,220&  28.34       & 34.86        &38.73
 \\\shline
\end{tabular}
\vspace{-20pt}
\end{table}
%%%%%%%%%%%%%%%%%%%%%%%%%%%%%%%%%%%%%%%%

\textbf{Effectiveness of CBS:} Settings (a) and (b) in Table~\ref{tab:ablation_study} give the PSNR comparison between w/ and  w/o  cross-block strategy (CBS). It is clear to see that  CBS greatly boosts the performance across all ratios, with the most significant improvement up to 0.64 dB, which fully verifies its effectiveness.

\vspace{-8pt}
\section{Conclusions}
\vspace{-6pt}
In this paper, we propose a flexible deep unfolding network named ISTA-Net$^{++}$ for CS reconstruction with strong flexibility and superior performance. By adopting a dynamic unfolding strategy and using a balanced training dataset, our method enjoys great flexibility to solve multi-ratio tasks and much robustness to reconstruct multi-scene images, thus being a suitable method for practical applications. Besides, a cross-block strategy is introduced to alleviate blocking artifacts and further improve the performance. Considering its flexibility, effectiveness and practicability, our model is expected to serve as a suitable baseline in future CS research.

\bibliographystyle{IEEEtran}
\small
\bibliography{icme2021template}

\end{document}